\documentclass[letterpaper]{article} % DO NOT CHANGE THIS
\usepackage[]{aaai25}  % DO NOT CHANGE THIS
\usepackage{times}  % DO NOT CHANGE THIS
\usepackage{helvet}  % DO NOT CHANGE THIS
\usepackage{courier}  % DO NOT CHANGE THIS
\usepackage[hyphens]{url}  % DO NOT CHANGE THIS
\usepackage{graphicx} % DO NOT CHANGE THIS
\usepackage{natbib}  % DO NOT CHANGE THIS AND DO NOT ADD ANY OPTIONS TO IT
\usepackage{caption} % DO NOT CHANGE THIS AND DO NOT ADD ANY OPTIONS TO IT
\usepackage{algorithm}
\usepackage{newfloat}
\usepackage{listings}
\DeclareCaptionStyle{ruled}{labelfont=normalfont,labelsep=colon,strut=off} % DO NOT CHANGE THIS
\floatstyle{ruled}
\newfloat{listing}{tb}{lst}{}
\floatname{listing}{Listing}
\usepackage{tabularx}
\usepackage{multirow} 
\usepackage{enumitem}
\usepackage{diagbox}
\usepackage{algorithm}
\usepackage{amssymb}
\usepackage{amsmath}
\usepackage[noend]{algpseudocode}
\usepackage{booktabs}
\usepackage{hyperref}

\usepackage{arydshln}

\usepackage{times}
\usepackage{latexsym}

\usepackage{enumitem}
\usepackage{amsfonts}
\usepackage{color}

\usepackage[utf8]{inputenc}

\usepackage{microtype}

\usepackage{inconsolata}

\newtheorem{theorem}{Theorem}[section]
\newtheorem{corollary}[theorem]{Corollary}
\newtheorem{lemma}[theorem]{Lemma}

\definecolor{commentcolor}{RGB}{110,154,155}   %
\newcommand{\PyComment}[1]{\ttfamily\textcolor{commentcolor}{\# #1}}  %
\newcommand{\PyCode}[1]{\ttfamily\textcolor{black}{#1}} %

\usepackage[table,xcdraw]{xcolor}

\title{Layer-wise Importance Matters: Less Memory for Better Performance in Parameter-efficient Fine-tuning of Large Language Models}

\author{
 \textbf{Kai Yao\textsuperscript{\rm 1,\rm 2}\thanks{These authors contributed equally to this work.}},
 \textbf{Penglei Gao\textsuperscript{\rm 3}\footnotemark[1]},
 \textbf{Lichun Li\textsuperscript{\rm 2}},
 \textbf{Yuan Zhao\textsuperscript{\rm 2}},
 \\
 \textbf{Xiaofeng Wang\textsuperscript{\rm 3}},
 \textbf{Wei Wang\textsuperscript{\rm 2}\thanks{Corresponding authors.}},
 \textbf{Jianke Zhu\textsuperscript{\rm 1}\footnotemark[2]},
\\
 \textsuperscript{1}Zhejiang University \ 
 \textsuperscript{2}Ant Group \ 
 \textsuperscript{3}Cleveland Clinic Lerner Research Institution
\\
\href{mailto:jiumo.yk@antgroup.com}{jiumo.yk@antgroup.com}, \href{mailto:gaop@ccf.org}{gaop@ccf.org}
}

\begin{document}
\maketitle
\begin{abstract}
Parameter-Efficient Fine-Tuning (PEFT) methods have gained significant popularity for adapting pre-trained Large Language Models (LLMs) to downstream tasks, primarily due to their potential to significantly reduce memory and computational overheads. However, a common limitation in most PEFT approaches is their application of a uniform architectural design across all layers. This uniformity involves identical trainable modules and ignores the varying importance of each layer, leading to sub-optimal fine-tuning results. To overcome the above limitation and obtain better performance, we develop a novel approach, Importance-aware Sparse Tuning (IST), to fully utilize the inherent sparsity and select the most important subset of full layers with effective layer-wise importance scoring. The proposed IST is a versatile and plug-and-play technique compatible with various PEFT methods that operate on a per-layer basis. By leveraging the estimated importance scores, IST dynamically updates these selected layers in PEFT modules, leading to reduced memory demands. We further provide theoretical proof of convergence and empirical evidence of superior performance to demonstrate the advantages of IST  over uniform updating strategies. Extensive experiments on a range of LLMs, PEFTs, and downstream tasks substantiate the effectiveness of our proposed method, showcasing IST's capacity to enhance existing layer-based PEFT methods. Our code is available at \url{https://github.com/Kaiseem/IST}%{https://github.com/Kaiseem/IST}}

\end{abstract}

\begin{figure}[!t]
\centering
\includegraphics[width=1\columnwidth]{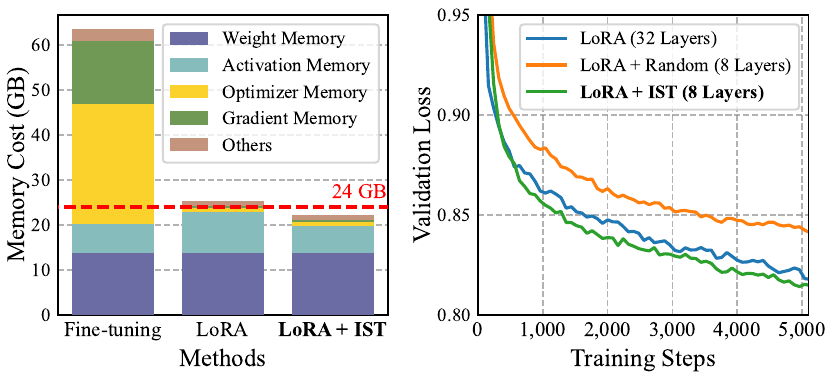}
\caption{(Left) Memory consumption of tuning a LLaMA 7B model with a token batch size of 1024 on a single device. Details refer to \autoref{sec:mem}. (Right) In comparison to the vanilla tuning of all 32 and random tuning of 8 LoRA layers, IST achieves a better validation loss.}
\label{fig:memory_breakdown} 
\end{figure}
\section{Introduction}
Significant achievements in natural language processing (NLP) have been achieved this year from the use of large language models (LLMs) that are pre-trained on extensive general datasets~\cite{zhuang2024toolqa,brown2020language}. These LLMs typically require full fine-tuning (FFT)~\cite{howard2018universal} to adapt them for specialized downstream tasks, an approach that necessitates retraining all model parameters. Nevertheless, as the size of these models and the volume of data increase, FFT becomes increasingly costly and impractical. Aiming to reduce the cost, parameter-efficient fine-tuning (PEFT) methods, involving  adapter-based~\cite{series,wang2022adamix,lei2024conditional,parallel}, reparameterization-based~\cite{lora,krona,dora}, and prompt-based methods~\cite{prefix,ptuning,prompttuning}, have been proposed to reduce the number of trainable parameters in fine-tuning for the downstream tasks. 
However, most existing PEFT methods employ a uniform approach that indiscriminately assigns trainable parameters to identical positions across all layers, which could be unnecessary. This strategy relies heavily on human heuristics and overlooks task-specific domain gaps and characteristics, limiting their performance across various downstream tasks.
Although some PEFT methods have improved the efficiency of fine-tuning LLMs, such as dynamic rank \cite{zhang2023adaptive,zhang2024autolora}, they are tailored specifically for LoRA-based models and do not extend their benefits to the additional learnable module-based methods, i.e., Series and Parallel configurations. This limitation creates a clear necessity for a more generalized algorithm to enhance model performance across various domains.

Inspired by LISA~\cite{pan2024lisa}, we empirically found that training a small fraction of the full layers in PEFT can yield comparably promising results to those achieved with FFT.
The existing PEFT methods exhibit markedly redundancy in layer updating during the training process, leading us to investigate the differences among the layers of varying importance from the perspective of layer-wise sparsity. Motivated by these inherent insights, we propose a novel PEFT-compatible plug-and-play approach, \textbf{I}mportance-aware \textbf{S}parse \textbf{T}uning (IST), that estimates the task-specific importance score of each layer and fine-tunes the most important ones. As shown in \autoref{fig:memory_breakdown}, our method substantially lowers memory demands during training by reducing the number of layers that require updates. Furthermore, by integrating layer-wise sparsity into our methodology, we enhance the convergence of layer-based PEFT methods, thereby achieving improved performance. The experimental results show that IST consistently improves existing layer-wise PEFT methods without sacrificing performance and inference efficiency across a wide range of models.

In summary, our contributions are as follows:
%Our contribution could be summarized into threefold:
\begin{itemize}[leftmargin=*]
\item Based on the empirical insight that sparse patterns markedly enhance the convergence of PEFT models, we propose an importance-aware sparse tuning method that prioritizes the most important layers for updating, making PEFT memory efficient and more powerful.
%. This focused strategy substantially decreases both memory and computational requirements. %We found that noticeable patterns of sparsity can markedly improve the rate of convergence for PEFT models and proposed an importance-aware sparse tuning method to selectively update the most significant layers. This targeted approach reduces memory and computational demands significantly.
\item We provide theoretical proof of convergence for the IST approach and present empirical evidence showing that it outperforms traditional uniform update strategies for PEFT.
%in terms of both performance and efficiency.
\item Extensive experiments in various LLMs, PEFT methods, and downstream tasks demonstrate the effectiveness and capacity of IST to enhance existing PEFT without sacrificing performance.
\end{itemize}

\section{Related Work}
 \subsection{Parameter-efficient Fine-tuning}
As models grow in size and complexity, pre-trained Large Language Models (LLMs) have shown impressive performance across a range of natural language processing (NLP) tasks. However, efficiently adapting these LLMs to specific downstream tasks poses increasing challenges. Parameter-efficient fine-tuning (PEFT) addresses this dilemma by fine-tuning a few additional parameters or a subset of the pre-trained parameters. The existing PEFT approaches can be roughly categorized into three main types: adapter-based~\cite{series,wang2022adamix,lei2024conditional,parallel}, reparameterization-based~\cite{lora,krona,dora}, and prompt-based methods~\cite{prefix,ptuning,prompttuning}. Adapter-based methods focus on adding extra tunable parameters by introducing new layers within the original model. For example, Series Adapters~\cite{series} incorporate linear modules in a sequential manner, whereas Parallel Adapters~\cite{parallel} add learnable modules in parallel with the model's existing sublayers. Meanwhile, reparameterization-based methods aim to reduce the total number of trainable parameters by employing low-rank representations. LoRA~\cite{lora}, a notably effective and popular method, breaks down the delta parameter matrix into two lower-rank matrices. 
Yet, most current PEFT methods apply a uniform architectural approach across all layers, utilizing the same trainable modules for each layer. In this study, we present a novel approach that dynamically tunes a subset of full layers through PEFT, significantly enhancing both training efficiency and the performance of the fine-tuned models.

\begin{figure*}[ht]
\centering
\includegraphics[width=2\columnwidth]{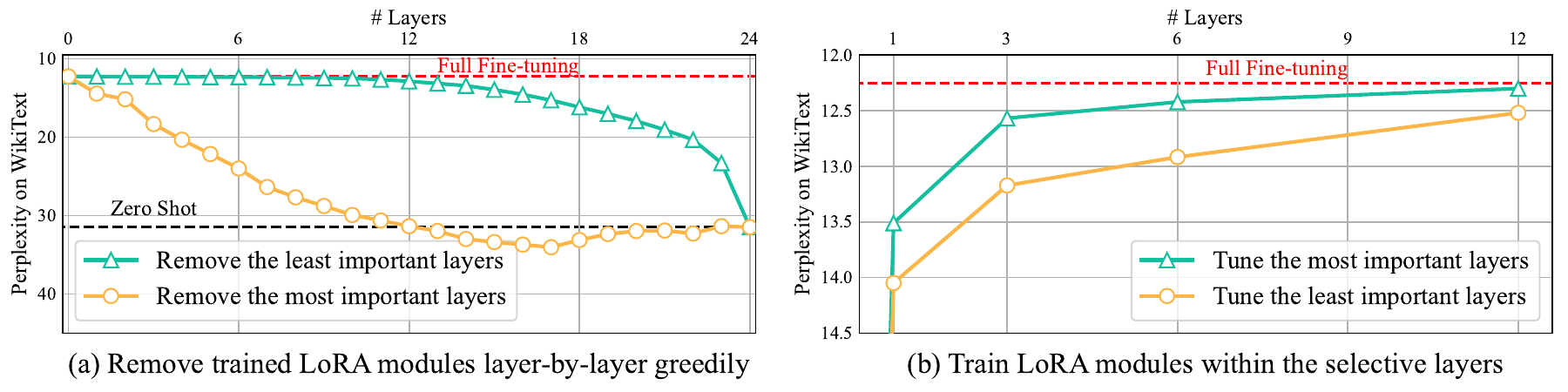}
\caption{Illustration of layer redundancy in PEFT training on the OPT-1.3B. (a) A greedy selection strategy is employed to iteratively remove the trained LoRA modules from the model. (b) Specific layers of the model are selectively fine-tuned using LoRA. The importance of layers depends on their contribution to the performance.}% #
\vspace{-2pt}
\label{fig:motivate} 
\vspace{-8pt}
\end{figure*}

\subsection{Layer-wise Sparse Tuning}
Many previous works have uncovered the phenomenon of layer redundancy in pre-trained models, as evidenced by methods such as LayerSkip~\cite{elhoushi2024layer}, LayerDrop~\cite{layerdrop}, LayerSharing~\cite{crash, layersharing}, and structured pruning~\cite{fan2021layer,zhang2020accelerating}. This indicates that the importance of each layer could be different, and not all layers need fine-tuning. However, selecting the appropriate layers for fine-tuning downstream tasks remains a significant challenge. \citet{lee2022surgical} suggests selectively fine-tuning a subset of layers depending on the type of domain shift. Similarly, \citet{kaplun2023less} deploys a greedy search to find the most suitable layers for fine-tuning, demanding considerable computational resources and time for initiation. Recently, layer-wise sparse training for large language models has become a popular topic. For example, LISA~\cite{pan2024lisa} randomly selects a subset of layers to be optimized during training, leading to promising faster convergence and better performance. LIFT~\cite{zhu2023lift} selects one layer to fine-tune LLMs with different selection strategies such as front-to-end, end-to-front, or random, obtaining comparable performance while reducing the computational load. Although effective, these methods require substantial storage equivalent to the full model since all parameters are being updated. Furthermore, these approaches do not deeply explore joint use with PEFT and have employed relatively simple selection strategies, limiting their performance. Unlike these previous methods, we focus on integrating existing layer-based PEFT and propose an importance-aware layer selection strategy that significantly enhances performance while increasing efficiency.

\section{Method}

\subsection{Motivation}
To showcase the excessive layer redundancy in training PEFT, we conducted empirical evaluations on the OPT 1.3B~\cite{opt}  model fine-tuned on the WikiText~\cite{wikitext} dataset.  Initially, we adopted LoRA on all model's layers and trained it on this dataset. After training, we employed a greedy selection strategy to remove the least or the most important layers individually according to their contribution to the model's performance. On the one hand, as illustrated in \autoref{fig:motivate}(a), removing 50\% of the least important LoRA layers did not substantially elevate perplexity. On the other hand, removing the most important LoRA layers resulted in a rapid decline in performance.
These preliminary findings indicate inherent layer-wise sparsity during PEFT training, leading to the phenomenon that not all layers are effectively trained with PEFT.
%highlighting significant redundancy within all the layers. Consequently, not all layers are effectively trained with PEFT, even when all are subjected to the training process. 

This observation prompts us to question: \textit{what causes the layer-wise sparsity?} To answer this question, we utilized the outcome of the greedy search to rank the layers according to their contribution to the model's performance. Next, we performed PEFT fine-tuning on the most and least important layers. As shown in \autoref{fig:motivate}(b), even when only a small portion of the layers (or merely a single one) are trained using LoRA, it is possible to attain comparable results to those obtained through full fine-tuning (FFT). This suggests that the observed sparsity is not due to the unimportant layers of the original network. Instead, it implies that the layer-wise sparsity observed in PEFT is an inherent characteristic, naturally emerging throughout the network's training process. Furthermore, training more important layers yields better outcomes consistently than focusing on the less important ones, emphasizing the beneficial role of importance in layer-wise sparsity.

\subsection{Convergence of Layer-wise Sparse Tuning}
%\yk{In the following, we will proof xxxx by xxxx}. 
In the following, we will demonstrate why layer-wise sparse tuning is efficient and effective during fine-tuning. In particular, we develop proof that if we randomly select a subset of full layers in the layer-based PEFT method and only update these selected parameters, the risk bond of the subsets can be tighter than updating the whole layers.

Given a pretrained Large Language Model (LLM) $\mathcal{M} = \{m_1,m_2,\dots,m_{N_L}\}$, comprising $N_L$ layers and parameterized by $\Theta$, alongside a downstream dataset $\mathcal{D}=\{(x_i,y_i)\}_{i\in{[ |\mathcal{D}|]}}$, full fine-tuning (FFT) this model on the downstream dataset achieve $\mathcal{M}_{\Theta} \rightarrow \mathcal{M}_{\Theta+\Delta}$, $\Delta=\arg\min_{\Delta}\mathcal{L}(\Theta+\Delta,\mathcal{D})$. PEFT introduces a learnable module $\mathcal{A}$ with a significantly smaller number of trainable parameters, denoted as $\mathcal{M}'=[\mathcal{M}_{\Theta}, \mathcal{A}]$, where $|\theta^{\mathcal{A}}| \ll |\Delta|$, aiming to achieve performance comparable to the fully fine-tuned model $\mathcal{M}_{\Theta+\Delta}$. The empirical loss over the training set $\mathcal{D}$ is defined as $\mathcal{L}(\theta^\mathcal{A};(x,y))=\frac{1}{|D|}\sum_{i\in{[ |\mathcal{D}|]}} \ell(y_i,f(x_i;\theta^\mathcal{A}))$, where $\ell$ denotes a suitable loss function, such as cross-entropy.

The learnable module $\mathcal{A}$ in most existing PEFT methods can be represented as $\mathcal{A} = \{a_1,a_2,\dots,a_{N_L}\}$. According to the sparse tuning strategy, the total layers of the given adapter module can be divided into two groups: $S$ and $\Bar{S}$ represent a set of randomly selected layers that are updated and a set of remaining layers that are frozen during fine-tuning respectively.
The total parameter vector $\theta^\mathcal{A}$ is then partitioned into $\theta^\mathcal{A}_S$ and $\theta^\mathcal{A}_{\Bar{S}}$. The loss function can conceptually be decomposed as follows:
\begin{equation}
    \mathcal{L}(\theta^\mathcal{A};(x,y))=\mathcal{L}(\theta^\mathcal{A}_S,\theta^\mathcal{A}_{\Bar{S}};(x,y)).
\end{equation}

\begin{corollary}
    Consider the Taylor expansion of the loss function $\mathcal{L}(\theta^\mathcal{A}_S,\theta^\mathcal{A}_{\Bar{S}})$ around $\theta^\mathcal{A}_{\Bar{S}}$:
    \begin{equation}
    \begin{split}
        \mathcal{L}(\theta^\mathcal{A}_S,\theta^\mathcal{A}_{\Bar{S}}) &= \mathcal{L}(\theta^\mathcal{A}_S,\theta^\mathcal{A}_{\Bar{S}^0})  \\
        & + \nabla_{\theta^\mathcal{A}_{\Bar{S}}}\mathcal{L}(\theta^\mathcal{A}_S,\theta^\mathcal{A}_{\Bar{S}^0})\top(\theta^\mathcal{A}_{\Bar{S}} - \theta^\mathcal{A}_{\Bar{S}^0}) \\
        & + \mathcal{O}((\theta^\mathcal{A}_{S})^2),
    \end{split}
    \label{taylor-l}
    \end{equation}
    where $\theta^\mathcal{A}_{\Bar{S}^0}$ represents the fixed parameters before any fine-tuning. Since $\theta^\mathcal{A}_{\Bar{S}}$ does not change during fine-tuning process, we can set $\theta^\mathcal{A}_{\Bar{S}}=\theta^\mathcal{A}_{\Bar{S}^0}$. The first-order term of Eq.~\ref{taylor-l} can be eliminated and we can have the approximate loss:
    \begin{equation}
        \mathcal{L}(\theta^\mathcal{A}_S,\theta^\mathcal{A}_{\Bar{S}}) \approx \mathcal{L}(\theta^\mathcal{A}_S,\theta^\mathcal{A}_{\Bar{S}^0}) \propto \mathcal{L}(\theta^\mathcal{A}_S).
        \label{coro3.3}
    \end{equation}
\end{corollary}
This estimation shows that the loss function mainly depends on the updates of $\theta^A_S$, supporting the decision to focus updates on the subset of full layers.

In Vapnik–Chervonenkis (VC) theory~\cite{devroye1996vapnik}, the VC-dimension denoted as $VCdim(\mathcal{H})$ is a measure of the size, i.e., capacity, complexity, expressive power, richness, or flexibility, of a class of sets $\mathcal{H}$. For neural networks, including LLMs, the VC-dimension typically increases with the number of trainable parameters.
Let $d_S$ be the VC-dimension of subset $\mathcal{H}_S$ and $d$ be the VC-dimension of full set $\mathcal{H}$. By updating only a subset of parameters $\theta^\mathcal{A}_S$, the effective VC-dimension $d_S$ of the hypothesis class corresponding to these parameters is reduced, which leads to a tighter generalization bound:
\begin{lemma}
    With a probability at least $1-\delta$ over the choice of a training set of size $n$, the following bound holds for $\mathcal{H}_S \subseteq \mathcal{H}$:
    \begin{equation}
        \begin{split}
            &|\mathcal{R}(\mathcal{H})-\hat{\mathcal{R}}_n(\mathcal{H})| \approx |\mathcal{R}(\mathcal{H}_S)-\hat{\mathcal{R}}_n(\mathcal{H}_S)| \\
            &\le \sqrt{\frac{Cd_S log(n/d_S) + log(1/\delta)}{n}},
        \end{split}
    \end{equation}
    where $\mathcal{R}(\mathcal{H}_S)=\mathbb{E}_{(x,y)\sim D}\mathcal{L}(\theta^\mathcal{A}_S;x,y)$ denotes the expected risk under the data distribution $D$ and $\hat{\mathcal{R}}_n(\mathcal{H}_S)=\frac{1}{n}\sum_{i=1}^n\mathcal{L}(\theta^\mathcal{A}_S;x_i,y_i)$ denotes the generalization risk on the specific dataset. $C$ is a constant related to the model and data distribution. Since $d_S \le d$, the generalization bound becomes tighter, implying that models with fewer updating layers generalize better assuming the same number of training samples.
\end{lemma}

Based on Eq.~\ref{coro3.3}, the generalization error of the model can be formally estimated as $|\mathcal{R}(\mathcal{H})-\hat{\mathcal{R}}_n(\mathcal{H})| \approx |\mathcal{R}(\mathcal{H}_S)-\hat{\mathcal{R}}_n(\mathcal{H}_S)|$.

When $\theta^\mathcal{A}_S$ is updated and $\theta^\mathcal{A}_{\Bar{S}}$ remains fixed, the model effectively reduces the dimensionality of the optimization problem. This can potentially lead to a more focused and efficient parameter search:
\begin{corollary}
    The derivative of $\mathcal{L}(\theta^\mathcal{A})$ with respect to $\theta^\mathcal{A}_S$ can be obtained as:
    \begin{equation}
        \frac{\partial \mathcal{L}(\theta^\mathcal{A})}{\partial \theta^\mathcal{A}_S} = \frac{1}{|D|}\sum_{i \in [|\mathcal{D}|]}\frac{\partial \ell((y_i,f(x_i;\theta^\mathcal{A})))}{\partial \theta^\mathcal{A}_S}.
    \end{equation}
    The magnitude and direction of this gradient tell us how sensitive the empirical risk is to changes in $\theta^\mathcal{A}_S$ and hence guide the updates during training.
\end{corollary}

From the above analysis, we found that noticeable patterns of sparsity combined with the smoothness of the objective function, can markedly improve the rate of convergence, potentially leading to a linear speed-up. To achieve improved error bounds and convergence rates, the crucial strategy lies in selecting the most important layers of the full model that are particularly pertinent to the specific task. This selection process involves identifying which layers contribute the most to task-specific performance, enabling a more focused and efficient training regimen.

\subsection{Importance-aware Sparse Tuning}
\begin{figure}[!t]
\centering
\includegraphics[width=1\columnwidth]{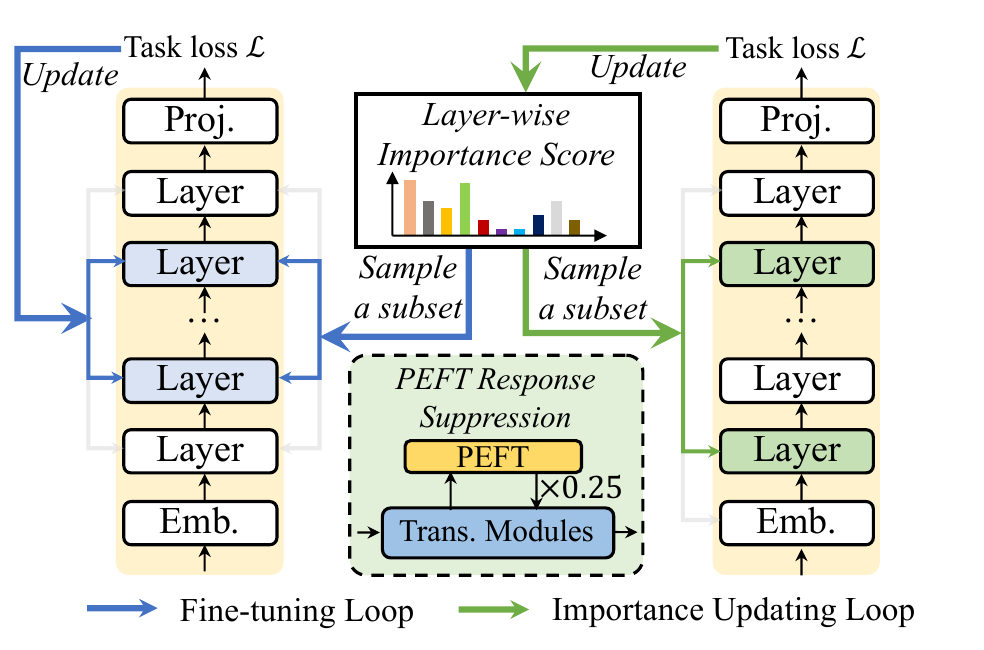}
\caption{Workflow of Importance-aware Sparse Tuning (IST): IST consists of two main loops: a fine-tuning loop, which selects a subset of layers for updating PEFT modules, and an importance updating loop, which estimates layer-wise importance by assessing the response suppression of the selected PEFT modules.}
\label{fig:ist} 
\vspace{-8pt}
\end{figure}

In the previous section, we proved that sparse tuning leads to a better convergence for downstream task fine-tuning. In this section, we will introduce our method, Importance-aware Sparse Tuning (IST), aiming to enhance the performance of layer-wise sparse tuning motivated by empirical observations. IST involves two loops: the fine-tuning loop, which selects a subset of full layers to update the PEFT modules, and the importance updating loop, which updates the importance score of each layer. To estimate layer-wise importance more accurately, we dynamically select the subsets of all layers for PEFT response suppression during the importance updating process. Drawing inspiration from reinforcement learning, which explores the best structure based on rewards~\cite{conf/iclr/ZophL17, conf/icml/PhamGZLD18, liu2017hierarchical}, we treat the layer selection process as a multi-armed bandit problem and use reinforcement learning to obtain the importance score of each layer.

\begin{table*}[t]
\setlength{\tabcolsep}{1.2mm}
\centering
\begin{tabular}{l|c|c|ccc|ccc}
\toprule
\multirow{2}{*}{\textbf{Model}} &\multirow{2}{*}{\textbf{Weight Mem.}} & \multirow{2}{*}{\textbf{Full Fint-tuning}} & \multicolumn{3}{c|}{\textbf{PEFT}} & \multicolumn{3}{c}{\textbf{PEFT + IST}} \\\cline{4-9} 
 & & & Series & Parallel & LoRA & Series & Parallel & LoRA \\\hline
GPT2-Small$_{\text{120M}}$ &0.4G& 3.4G & 2.5G & 2.8G & 2.9G & 2.0G & 2.1G & 2.2G \\
TinyLLaMA$_{\text{1.1B}}$ &2.2G& 15.9G & 9.6G & 10.3G & 10.5G & 8.0G & 8.3G & 8.5G \\
LLaMA$_{\text{7B}}$ &14G& 60G & 22G & 23G & 24G & 19G & 20G & 20G \\
LLaMA$_{\text{13B}}$ &27G& OOM & 38G & 41G & 42G & 35G & 36G& 36G \\\bottomrule
\end{tabular}
\vspace{-2pt}
\caption{Comparison of memory consumption for various LLMs and PEFT methods.}
\label{memory}
\vspace{-8pt}
\end{table*}

\paragraph{Fine-tuning Loop} Formally, given a PEFT-equipped LLM for downstream data fine-tuning, $\mathcal{M}'=[\mathcal{M},\mathcal{A}]=\{m_i, a_i\}_{i=1}^{N_L}$, where $m_i$ is frozen and $a_i$ is trainable,
our goal is to generate a subset $S$ of full layers to update, and keep the remaining set $\Bar{S}$ unchanged. To this end, we first define the degree of importance as $\mathbf{I}\in \mathbb{R}^{N_L}$, which is zero-initialized and updated through fine-tuning process simultaneously.
In each training iteration, we choose $N_u$ layers to update based on $\mathbf{I}$. For $t$-th step, the action policy $\pi_i$ for $i$-th layer follows the uniform distribution:
\begin{equation}
    \pi_i \sim U(0,\text{Sigmoid}(\mathbf{I}_i)).
\end{equation}
%where $U(a,b)$ is the uniform distribution between $a$ and $b$.
We randomly sample probability score $p_i$ for each layer, i.e., $p_i \sim \pi_i$. %We sample the probability score $p_i$ for each layer as $p_i \sim \pi_i$.
The subset $S$ can be determined with the score $p_i$:
\begin{equation}
    S =\{i| p_i> p_{N_u}\}, \Bar{S} =\{i| p_i \le p_{N_u}\},
\end{equation}
where $p_{N_u}$ is the $N_u$ largest values in the sampled probabilities. Then, the chosen PEFT modules are updated by $\theta_{a_i, i\in S}\leftarrow \nabla_{\theta^\mathcal{A}}\mathcal{L}(\theta^\mathcal{A})$

\paragraph{Importance Updating Loop} To update the importance score, we suppress the response of $a_i$ to measure its contribution to the result. If $a_i$ is relatively important, reducing its response will significantly increase the loss, and vice versa. %To accurately estimate the importance score, 
We sample $N_c$ candidate sets $\{S_c^1,\dots,S_c^{N_c}\}$, each containing $N_{v}$ layers. For the $j$-th sampling, we reduce the response of $a_i$ for the layers that were not selected:
\begin{equation}
\small
    o_{i+1}^j = \left\{\begin{array}{ll}
m_i(o_i^j)+a_i(o_i^j) & \text{if } i \in S_c^j \\
m_i(o_i^j)+\beta * a_i(o_i^j) & \text{otherwise}
\end{array} \right. ,
\end{equation}
where $\beta \in [0, 1]$ is the response suppression factor. Then, for the $j$-th sampled set $S_c^j$, we calculate the rewards according to their loss:
\begin{equation}
\small
\textbf{r}^j = e^{-\mathcal{L}^j}- \frac{1}{N_c} \sum\nolimits_{k=1}^{N_c} e^{-\mathcal{L}_k}.
\end{equation}
Due to the smaller contributions of PEFT compared to the original network, the response suppression of PEFT may lead to relatively small reward values. Therefore, we employed a large updating rate $\mu$ to accelerate the convergence of importance, ensuring it matches the fine-tuning process:
\begin{equation}
\small
\label{eq:6}
\mathbf{I}_i = \left\{
\begin{array}{ll}
\mathbf{I}_i + \mu*\mathbf{r}_j & \text{if } i \in S_c^j \\
\mathbf{I}_i & \text{otherwise}
\end{array}
\right. ,
\end{equation}
where $\mu$ controls the convergence of importance.

\paragraph{Joint Training} 
We propose jointly training IST with PEFT to avoid the costly greedy search observed in prior studies~\cite{kaplun2023less}, as shown in \autoref{fig:ist}. Specifically, to align with the training dynamics of PEFT, we execute the importance updating loop every $T_c$ fine-tuning loop. While our method reduces the time required for the fine-tuning loop slightly, it introduces additional forward time within the importance updating loop. Consequently, we set $T_c=10$ and $N_c=3$ to keep the training time efficient.

\section{Experimental Results}
In this section, we conduct a series of experiments to validate the effectiveness of our proposed IST. We integrate IST into the Series Adapter, Parallel Adapter, and LoRA, and then compare them with their original counterparts across various tasks. 

\paragraph{Baselines} We included the following widely used layer-based fine-tuning methods.
\begin{itemize}[leftmargin=*]
\item \textbf{Full Fine-tuning}~\cite{howard2018universal} - All parameters within the pre-trained model are optimized during training.
\item \textbf{Series Adapter}~\cite{series} - Additional learnable modules are introduced into a specific sublayer in a sequential manner. 
\item \textbf{Parallel Adapter}~\cite{parallel} - Additional learnable modules are integrated in parallel with distinct sublayers within the backbone model.
\item \textbf{LoRA}~\cite{lora} - Parameter efficiency is enhanced by decomposing the learnable delta parameter matrix into two low-rank matrices.
\end{itemize}

For the optimal configuration and placement of PEFT methods, we adhere to the settings established by \citet{llmadapter}. Specifically, Series and Parallel Adapters are seamlessly integrated into the MLP layers with a bottleneck size of 256. Similarly, LoRA is seamlessly incorporated into both the Multi-head Self-attention layers and the MLP layers, with a rank of 32. Across all PEFT methods, we maintain the same tunable parameter budgets, adjusting only the learning rate. For IST, we consistently set $N_u$ to 25\% of the layers for the fine-tuning loop, $N_v$ to 50\% of the layers for the importance updating loop, and $\beta$ to 0.25. Further details on the experimental settings are available in the Appendix.

\subsection{Memory Efficiency}
\label{sec:mem}
We conducted experiments on maximum GPU memory to demonstrate the efficiency of IST in terms of memory usage, revealing that it requires less memory compared to standard PEFT methods.

\noindent\paragraph{Settings} To obtain an accurate estimation of the memory, we randomly sampled prompts from the Alpaca~\cite{alpaca} dataset and restricted the maximum output token length to 1024. We uniformly employed a mini-batch size of 1 across four LLMs, ranging from 120M to 13B parameters, and three types of PEFT methods. We presented the overall memory consumption, consisting of weight memory, activation memory, optimizer memory, and gradient memory. Additionally, we separately demonstrated weight memory to highlight the significant role of IST in reducing training memory. To isolate the impact of the evaluated variables, we excluded GPU memory-saving techniques, such as gradient checkpointing~\cite{gradientcheckpointing}, offloading~\cite{offloading}, and flash attention~\cite{dao2022flashattention}.

\begin{table*}[t]
% \vspace{0.8em}
\setlength{\tabcolsep}{1.2mm}
\centering
\resizebox{0.99\textwidth}{!}{
\begin{tabular}{clcccccccccc}
\toprule
\textbf{Model} & \multicolumn{1}{c}{\textbf{PEFT}}  & \textbf{BoolQ} & \textbf{PIQA}&\textbf{SIQA}& \textbf{HellaSwag} & \textbf{WinoGrande}& \textbf{ARC-e} & \textbf{ARC-c} & \textbf{OBQA} & \textbf{Avg.} \\ \hline
ChatGPT&\multicolumn{1}{c}{-}&73.1&85.4&68.5&78.5&66.1&89.8&79.9&74.8&77.0\\\hline
\multirow{6}{*}{LLaMA$_{\text{7B}}$}&Series&63.0&79.2&76.3&67.9&75.7&74.5&57.1&72.4&70.8\\
&Series + IST&66.2&78.3&74.9&72.2&75.9&75.8&59.0&72.2&\textbf{71.8}\\%2e-4
&Parallel&67.9&76.4&78.8&69.8&78.9&73.7&57.3&75.2&72.2\\
&Parallel + IST&68.4&79.1&77.9&70.0&78.9&81.2&62.3&77.6&\textbf{74.4}\\%2e-4
&LoRA&68.9&80.7&77.4&78.1&78.8&77.8&61.3&74.8&74.7\\
&LoRA + IST&68.7&81.7&77.3&82.7&78.7&80.6&62.4&80.0&\textbf{76.5}\\\hline%2e-4
\multirow{6}{*}{LLaMA$_{\text{13B}}$}&Series&71.8&83.0&79.2&88.1&82.4&82.5&67.3&81.8&79.5\\
&Series + IST&  72.9&82.2&81.4&87.9&84.0&82.7&69.1&81.1&\textbf{80.2}\\%2e-4
%&Series + IST&  70.0&79.3&78.5&79.9&81.1&79.8&66.2&78.2&76.6\\%2e-4
%&Series + IST&  & & & & & & & & \\%2e-4
&Parallel&72.5&84.9&79.8&92.1&84.7&84.2&71.2&82.4&\textbf{81.4}\\
&Parallel + IST&72.6&86.0&79.2&89.1&83.5&84.8&70.6&82.8&81.1\\%2e-4,bottleneck size 128
%&Parallel + IST&&&&&&&&&\\%2e-4
&LoRA&72.1&83.5&80.5&90.5&83.7&82.8&68.3&82.4&80.5\\
&LoRA + IST&71.5&85.0&81.2&89.1&84.2&84.0&70.1&81.8&\textbf{80.9}\\\hline\hline%2e-4
\multirow{2}{*}{GPT-J$_{\text{6B}}$}&LoRA&62.4&68.6&49.5&43.1&57.3&43.4&31.0&46.6&50.2\\
&LoRA + IST&63.0&63.2&62.9&35.8&39.1&56.8&39.1&51.2&\textbf{51.4}\\\hline
\multirow{2}{*}{BLOOMz$_{\text{7B}}$}&LoRA&65.9&75.3&74.5&57.3&72.5&74.6&57.8&73.4&\textbf{68.9}\\
&LoRA + IST&67.0&74.4&74.4&51.4&68.7&77.9&58.9&74.4&68.4\\\hline%2e-4
\multirow{2}{*}{LLaMA3$_{\text{8B}}$}&LoRA&70.8&85.2&79.9&91.7&84.3&84.2&71.2&79.0&80.8\\
&LoRA + IST&72.7&88.3&80.5&94.7&84.4&89.8&79.9&86.6&\textbf{84.6}\\%1e-4
\bottomrule
% &DoRA&74.6&89.3&79.9&95.5&85.6&90.5&80.4&85.8&\textbf{85.2}\\
% &IST+DoRA&74.0&89.2&80.2&95.0&86.2&90.3&81.2&85.6&\textbf{85.2}\\
\end{tabular}}
\vspace{-2pt}
\caption{Accuracy comparison of multiple LLMs with various PEFT methods on eight commonsense reasoning datasets. Results of all the baseline methods on GPT-J, BLOOMZ and LLaMA are taken from \citet{llmadapter}.
}
\label{tab:llama_commonsense}
\vspace{-8pt}
\end{table*}

\noindent\paragraph{Results}
We list the memory consumption for various LLMs and PEFT methods in \autoref{memory}. The overall results show that training LLMs with our proposed IST strategy could significantly reduce memory consumption in all the widely used LLMs compared to full fine-tuning and standalone PEFT configurations. 
Combining PEFT modules and our proposed IST could save a lot of training memory including activation memory, optimizer memory, and gradient memory on the three popular adapters.
As for the LLaMA 7B model, training with IST could almost reduce the average $36\%$ training memory for all the PEFT methods. This trend of reduced memory usage with IST integration is consistent across other models as well.
These results highlight the effectiveness of IST in enhancing the memory efficiency of fine-tuning LLMs, which makes IST a valuable strategy in deploying more resource-efficient fine-tuning practices, especially important for scenarios where computational resources are a limiting factor. %This becomes particularly relevant in handling larger models like LLaMA13B, where traditional fine-tuning approaches reach operational limits. 

\subsection{Commonsense Reasoning}

\noindent\paragraph{Settings} To validate the effectiveness of IST, we evaluated three PEFT methods across five LLMs on the commonsense reasoning tasks. Specifically, the adaptability of PEFT was verified using Series, Parallel Adapter, and LoRA methods on the LLaMA 7/13B~\cite{llama} models, and the adaptability of LLM was tested on three models: GPT-J 6B~\cite{gpt-j}, BLOOMZ 7B~\cite{bloom}, and LLaMA3 8B~\cite{llama3modelcard}. 
We also report ChatGPT’s accuracy obtained with gpt-3.5-turbo API using a zero-shot Chain of Thought~\cite{wei2022chain}. 
The commonsense reasoning tasks consisted of 8 sub-tasks, each with a predefined training and testing set, including BoolQ~\cite{boolq}, PIQA~\cite{piqa}, SIQA~\cite{siqa}, HellaSwag~\cite{hellaswag}, WinoGrande~\cite{winogrande}, ARC~\cite{arc} and OBQA~\cite{openbookqa}. 
Aligning with the setting of \citet{llmadapter}, we aggregated the training data from all eight tasks to form the training dataset and conducted evaluations on the individual testing dataset for each task.

\noindent\paragraph{Results}
The quantitative results in \autoref{tab:llama_commonsense} offer a comprehensive view of the performance improvements brought by the proposed IST method across various LLMs and PEFT configurations. We can see that IST consistently shows an enhancement in model performance on the commonsense reasoning task. Analyzed from the LLaMA 7B model, IST shows significant performance gains across multiple tasks compared to its PEFT-only counterparts in all three PEFT configurations. Notably, in tasks HellaSwag and QBQA, there's a noticeable improvement, demonstrating how IST can refine the model’s response to more complex queries. Moreover, the impact of IST is not limited to one model or configuration. For example, in GPT-J 6B and BLOOMZ 7B, the IST enhancements lead to better outcomes in almost all tasks compared to LoRA configurations without IST. This across-the-board improvement underscores IST’s robustness and general applicability. IST’s ability to focus on the most impactful layers makes the fine-tuning process not only more memory efficient but also strategically adaptable to various reasoning tasks. This is particularly beneficial in scenarios where model responsiveness and accuracy are critical. The aggregation of training data across different tasks and the subsequent application of IST likely helps in developing a more generalized understanding of commonsense reasoning, making IST a valuable addition to the PEFT techniques.

\begin{table}[t]
\small
% \vspace{0.8em}
\setlength{\tabcolsep}{0.7mm}
\centering
\begin{tabular}{lcccccc}
\toprule
 \multicolumn{1}{c}{\textbf{Method}}  & \textbf{GSM8K} & \textbf{AQuA}&\textbf{MAWPS}& \textbf{SVAMP}& \textbf{Avg.} \\ \hline
\multicolumn{1}{c}{ChatGPT}&56.4&38.9&87.4&69.9&63.2\\\hline
LoRA&61.0&26.4&91.6&74.4&63.4\\
LoRA + IST&62.8&31.5&89.9&76.3&\textbf{64.7} \\
\bottomrule
\end{tabular}
\vspace{-2pt}
\caption{Accuracy comparison of LLaMA3 8B on four math reasoning datasets.}
\label{tab:llama_math}
\vspace{-8pt}
\end{table}

\subsection{Arithmetic Reasoning}

\noindent\paragraph{Settings}
To further demonstrate IST’s scalability on different tasks, we conduct additional fine-tuning experiments on arithmetic reasoning. We utilized LoRA to fine-tune the LLaMA 3 8B model. Similarly, we included the results from ChatGPT 3.5 as a reference, obtained using Zero-shot Chain-of-Thought~\cite{wei2022chain}. The fine-tuning process was conducted on the Math10K dataset, comprising math reasoning samples collected by \citet{llmadapter}. Following the completion of training, we evaluated the model's performance on predefined test sets from several datasets, including GSM8K~\cite{gsm8k}, AQuA~\cite{aqua}, MAWPS~\cite{mawps}, and SVAMP~\cite{svamp}.

\noindent\paragraph{Results}
\autoref{tab:llama_math} shows the results of the arithmetic reasoning task. The accuracy on GSM8K and SVAMP datasets shows a consistent improvement from ChatGPT to LoRA, and further enhancement when IST is applied alongside LoRA, indicating the effectiveness of fine-tuning and IST in improving model performance for these datasets. The results of the AQuA dataset indicate a decrease in accuracy for LoRA compared to ChatGPT, but the application of IST helps to recover some of the lost performance. This suggests that while LoRA alone may not be as effective for AQuA, IST can mitigate some issues. Overall, combining IST and existing PEFT methods presents a robust approach for fine-tuning LLMs, leading to better generalization and accuracy in arithmetic reasoning tasks.

\subsection{Analytical Study}
\paragraph{Effect of Importance-aware Sparse Tuning}
\begin{table}[]
\small
\setlength{\tabcolsep}{0.7mm}
\centering
\begin{tabular}{ccc}\toprule
Method & \# Layers & Results \\\midrule
Vanilla Tuning & 32 & 74.7 \\
Random Sparse Tuning & 4 & 67.1 (-7.6) \\
Importance-aware Sparse Tuning & 4 & 73.7 (-1.0)\\
Random Sparse Tuning & 8 & 75.8 (+1.1) \\
Importance-aware Sparse Tuning & 8 & 76.5 (+1.8)\\
\bottomrule
\end{tabular}
\caption{Ablation studies on key components of IST.}
\label{tab:ablate}
\end{table}

\begin{table}[]
\small
\setlength{\tabcolsep}{0.7mm}
\centering
\begin{tabular}{cccccc}\toprule
Method  & LoRA & LISA & AdaLoRA & \begin{tabular}[c]{@{}c@{}}LoRA\\ +IST\end{tabular} & \begin{tabular}[c]{@{}c@{}}AdaLORA\\ +IST\end{tabular} \\\midrule
Results & 74.7 & 75.3 & 76.2    & 76.5                                                & 77.1        \\
\bottomrule
\end{tabular}
\caption{Comparison with other adaptive methods.}
\label{tab:adaptivemethods}
\end{table}

We conducted experiments to evaluate the effects of importance-aware sparse tuning by training the LLaMA 7B model with LoRA on a commonsense task, reporting the average accuracy. As shown in \autoref{tab:ablate}, using sparse tuning with randomly selected layers, particularly with only four layers, does not yield satisfactory results. This outcome contrasts with findings from LISA~\cite{pan2024lisa} and LIFT~\cite{zhu2023lift}, where training very few layers (1-2 layers) resulted in a good performance. The discrepancy arises because, in LISA and LIFT, training entire transformer layers encompasses a substantial number of trainable parameters. Conversely, PEFT involves relatively fewer parameters, necessitating the fine-tuning of more layers to achieve better results. When we increased the number of sparse tuning layers to 8, we observed a considerable improvement of 1.1, aligning with theoretical expectations that sparse tuning enhances convergence. Finally, incorporating importance-aware tuning yielded the best results, underscoring the effectiveness of IST.

\begin{figure}[!t]
\centering
\includegraphics[width=1\columnwidth]{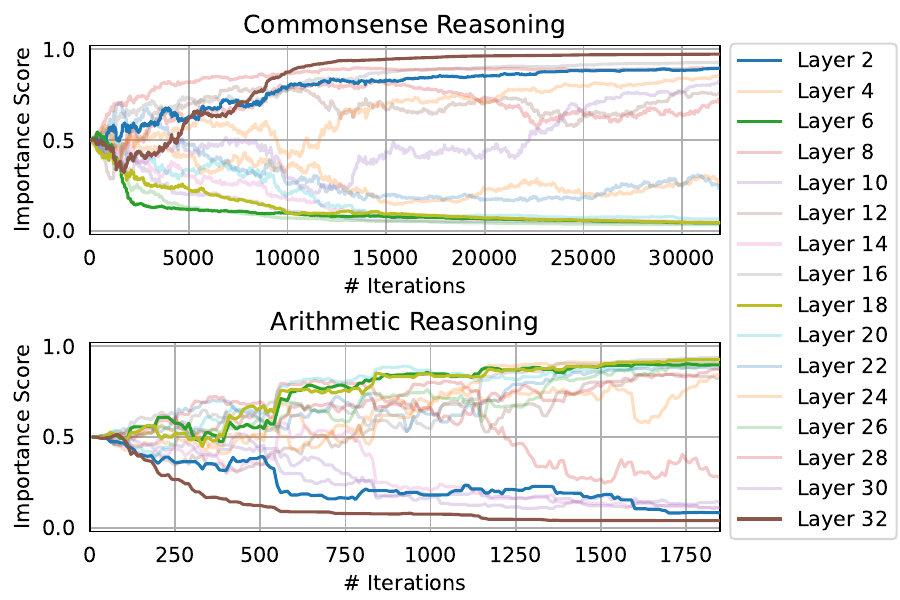}
\caption{Layer-wise importance on different tasks.}
\label{fig:task_imp} 
\vspace{-8pt}
\end{figure}
\paragraph{Comparison with Adaptive Methods}
We compared our proposed IST method with other adaptive methods, such as LISA~\cite{pan2024lisa} and AdaLoRA~\cite{zhang2023adaptive}, with LLaMa 7B on the commonsense task to demonstrate our effectiveness. Notably, LISA is a PEFT method that focuses on sparsely tuning a single transformer layer, while AdaLoRA uses adaptive rank allocation and can widely adapt to reparameterization-based methods. As shown in the Table~\ref{tab:adaptivemethods}, compared to LoRA, LISA improved the average accuracy by 0.6, validating the concept of sparse training. AdaLoRA improved accuracy by 1.5, highlighting the importance of rank-level sparsity. Finally, our method can be combined with LoRA and AdaLoRA to further enhance performance, showcasing the broad applicability and practicality of IST.

\paragraph{Layer-wise Importance Learning} We visualize the layer-wise importance of the two tasks with IST in \autoref{fig:task_imp}. The importance scores converge as the training iterations increase. The observed variation in the importance scores of each layer across different tasks indicates distinct levels of significance. %for each layer depending on the specific task at hand.
For instance, `Layer 2' and `Layer 32' significantly contribute to the commonsense reasoning task, whereas they are less important for the arithmetic reasoning task. Conversely, `Layer 6' and `Layer 18' exhibit contrasting importance levels across these tasks as well.
This layer-wise differentiation underscores the effectiveness of our method, similar to curriculum learning, where the model progressively focuses on the most pertinent layers at each stage of training. By dynamically adjusting the importance of different layers, our approach allows for a more refined and task-specific tuning process, thereby enhancing the model's adaptability and performance across diverse tasks.

\section{Conclusion}
In this study, we proposed a novel Importance-aware Sparse Tuning (IST) approach for PEFT of LLMs. By dynamically selecting the most important layers in the fine-tuning loop, IST achieves a significant reduction in memory usage and computational overhead. The importance updating loop refines the selection of layers using a reinforcement learning approach, ensuring that the most impactful layers are prioritized during training. This innovative method leverages the inherent sparsity of layer-wise importance, leading to more efficient and effective fine-tuning through extensive experiments across various LLMs, PEFT methods, and downstream tasks. The proposed method holds promise for future applications where resource constraints and performance are critical considerations.

\section*{Acknowledgements}
This work was supported by Ant Group Postdoctoral Programme.

\section*{Limitations}
There are three limitations in this work. First, since IST employs reinforcement learning, tuning six related hyperparameters is required. Even after fixing three of these parameters, the search space for the remaining three remains large, possibly leading to increased trial-and-error costs during usage. Second, due to limited resources, we were unable to validate larger language models such as the LLaMA3 70B. These larger models exhibit stronger language comprehension capabilities and, consequently, yield better performance. Third, we did not thoroughly explore the variants or combinations of each PEFT method. Given the substantial computational demands and extensive hyperparameter search space, we leave this as future work.

\bibliography{custom}

\begin{thebibliography}{60}
\providecommand{\natexlab}[1]{#1}

\bibitem[{AI@Meta(2024)}]{llama3modelcard}
AI@Meta. 2024.
\newblock Llama 3 Model Card.

\bibitem[{Bisk et~al.(2020)Bisk, Zellers, Bras, Gao, and Choi}]{piqa}
Bisk, Y.; Zellers, R.; Bras, R.~L.; Gao, J.; and Choi, Y. 2020.
\newblock PIQA: Reasoning about Physical Commonsense in Natural Language.
\newblock In \emph{Thirty-Fourth AAAI Conference on Artificial Intelligence}.

\bibitem[{Brown et~al.(2020)Brown, Mann, Ryder, Subbiah, Kaplan, Dhariwal, Neelakantan, Shyam, Sastry, Askell et~al.}]{brown2020language}
Brown, T.; Mann, B.; Ryder, N.; Subbiah, M.; Kaplan, J.~D.; Dhariwal, P.; Neelakantan, A.; Shyam, P.; Sastry, G.; Askell, A.; et~al. 2020.
\newblock Language models are few-shot learners.
\newblock \emph{Advances in neural information processing systems}, 33: 1877--1901.

\bibitem[{Chen et~al.(2023)Chen, Zhang, Shi, Li, Smola, and Yang}]{s4_model}
Chen, J.; Zhang, A.; Shi, X.; Li, M.; Smola, A.; and Yang, D. 2023.
\newblock Parameter-Efficient Fine-Tuning Design Spaces.
\newblock \emph{arXiv preprint arXiv:2301.01821}.

\bibitem[{Chen et~al.(2016)Chen, Xu, Zhang, and Guestrin}]{gradientcheckpointing}
Chen, T.; Xu, B.; Zhang, C.; and Guestrin, C. 2016.
\newblock Training deep nets with sublinear memory cost.
\newblock \emph{arXiv preprint arXiv:1604.06174}.

\bibitem[{Clark et~al.(2019)Clark, Lee, Chang, Kwiatkowski, Collins, and Toutanova}]{boolq}
Clark, C.; Lee, K.; Chang, M.-W.; Kwiatkowski, T.; Collins, M.; and Toutanova, K. 2019.
\newblock {B}ool{Q}: Exploring the Surprising Difficulty of Natural Yes/No Questions.
\newblock In \emph{Proceedings of the 2019 Conference of the North {A}merican Chapter of the Association for Computational Linguistics: Human Language Technologies, Volume 1 (Long and Short Papers)}, 2924--2936. Minneapolis, Minnesota: Association for Computational Linguistics.

\bibitem[{Clark et~al.(2018)Clark, Cowhey, Etzioni, Khot, Sabharwal, Schoenick, and Tafjord}]{arc}
Clark, P.; Cowhey, I.; Etzioni, O.; Khot, T.; Sabharwal, A.; Schoenick, C.; and Tafjord, O. 2018.
\newblock Think you have Solved Question Answering? Try ARC, the AI2 Reasoning Challenge.
\newblock \emph{arXiv:1803.05457v1}.

\bibitem[{Cobbe et~al.(2021)Cobbe, Kosaraju, Bavarian, Hilton, Nakano, Hesse, and Schulman}]{gsm8k}
Cobbe, K.; Kosaraju, V.; Bavarian, M.; Hilton, J.; Nakano, R.; Hesse, C.; and Schulman, J. 2021.
\newblock Training verifiers to solve math word problems.
\newblock \emph{arXiv preprint arXiv:2110.14168}.

\bibitem[{Dao et~al.(2022)Dao, Fu, Ermon, Rudra, and R{\'e}}]{dao2022flashattention}
Dao, T.; Fu, D.; Ermon, S.; Rudra, A.; and R{\'e}, C. 2022.
\newblock Flashattention: Fast and memory-efficient exact attention with io-awareness.
\newblock \emph{Advances in Neural Information Processing Systems}, 35: 16344--16359.

\bibitem[{Devroye et~al.(1996)Devroye, Gy{\"o}rfi, Lugosi, Devroye, Gy{\"o}rfi, and Lugosi}]{devroye1996vapnik}
Devroye, L.; Gy{\"o}rfi, L.; Lugosi, G.; Devroye, L.; Gy{\"o}rfi, L.; and Lugosi, G. 1996.
\newblock Vapnik-Chervonenkis Theory.
\newblock \emph{A probabilistic theory of pattern recognition}, 187--213.

\bibitem[{Edalati et~al.(2022)Edalati, Tahaei, Kobyzev, Nia, Clark, and Rezagholizadeh}]{krona}
Edalati, A.; Tahaei, M.~S.; Kobyzev, I.; Nia, V.; Clark, J.~J.; and Rezagholizadeh, M. 2022.
\newblock KronA: Parameter Efficient Tuning with Kronecker Adapter.
\newblock \emph{ArXiv}, abs/2212.10650.

\bibitem[{Elhoushi et~al.(2024)Elhoushi, Shrivastava, Liskovich, Hosmer, Wasti, Lai, Mahmoud, Acun, Agarwal, Roman et~al.}]{elhoushi2024layer}
Elhoushi, M.; Shrivastava, A.; Liskovich, D.; Hosmer, B.; Wasti, B.; Lai, L.; Mahmoud, A.; Acun, B.; Agarwal, S.; Roman, A.; et~al. 2024.
\newblock Layer skip: Enabling early exit inference and self-speculative decoding.
\newblock \emph{arXiv preprint arXiv:2404.16710}.

\bibitem[{Fan et~al.(2021)Fan, Li, Ao, Wu, Meng, and Sun}]{fan2021layer}
Fan, C.; Li, J.; Ao, X.; Wu, F.; Meng, Y.; and Sun, X. 2021.
\newblock Layer-wise model pruning based on mutual information.
\newblock \emph{arXiv preprint arXiv:2108.12594}.

\bibitem[{Fu et~al.(2021)Fu, Huang, Chen, Tian, and Zhao}]{lets}
Fu, C.; Huang, H.; Chen, X.; Tian, Y.; and Zhao, J. 2021.
\newblock Learn-to-Share: A Hardware-friendly Transfer Learning Framework Exploiting Computation and Parameter Sharing.
\newblock In Meila, M.; and Zhang, T., eds., \emph{Proceedings of the 38th International Conference on Machine Learning}, volume 139 of \emph{Proceedings of Machine Learning Research}, 3469--3479. PMLR.

\bibitem[{Han et~al.(2024)Han, Gao, Liu, Zhang, and Zhang}]{peft_survey}
Han, Z.; Gao, C.; Liu, J.; Zhang, J.; and Zhang, S.~Q. 2024.
\newblock Parameter-efficient fine-tuning for large models: A comprehensive survey.
\newblock \emph{ArXiv}, abs/2403.14608.

\bibitem[{He et~al.(2021)He, Zhou, Ma, Berg-Kirkpatrick, and Neubig}]{mam}
He, J.; Zhou, C.; Ma, X.; Berg-Kirkpatrick, T.; and Neubig, G. 2021.
\newblock Towards a unified view of parameter-efficient transfer learning.
\newblock \emph{arXiv preprint arXiv:2110.04366}.

\bibitem[{He et~al.(2022{\natexlab{a}})He, Zhou, Ma, Berg-Kirkpatrick, and Neubig}]{parallel}
He, J.; Zhou, C.; Ma, X.; Berg-Kirkpatrick, T.; and Neubig, G. 2022{\natexlab{a}}.
\newblock Towards a unified view of parameter-efficient transfer learning.
\newblock In \emph{International Conference on Learning Representations}.

\bibitem[{He et~al.(2022{\natexlab{b}})He, Ding, Dong, Zhang, and Tao}]{sparseadapter}
He, S.; Ding, L.; Dong, D.; Zhang, J.; and Tao, D. 2022{\natexlab{b}}.
\newblock {S}parse{A}dapter: An Easy Approach for Improving the Parameter-Efficiency of Adapters.
\newblock In \emph{Findings of the Association for Computational Linguistics: EMNLP 2022}, 2184--2190. Abu Dhabi, United Arab Emirates: Association for Computational Linguistics.

\bibitem[{Henderson, Ruder et~al.(2021)}]{compacter}
Henderson, J.; Ruder, S.; et~al. 2021.
\newblock Compacter: Efficient low-rank hypercomplex adapter layers.
\newblock In \emph{Advances in Neural Information Processing Systems}.

\bibitem[{Houlsby et~al.(2019)Houlsby, Giurgiu, Jastrzebski, Morrone, De~Laroussilhe, Gesmundo, Attariyan, and Gelly}]{series}
Houlsby, N.; Giurgiu, A.; Jastrzebski, S.; Morrone, B.; De~Laroussilhe, Q.; Gesmundo, A.; Attariyan, M.; and Gelly, S. 2019.
\newblock Parameter-efficient transfer learning for NLP.
\newblock In \emph{International conference on machine learning}, 2790--2799.

\bibitem[{Howard and Ruder(2018)}]{howard2018universal}
Howard, J.; and Ruder, S. 2018.
\newblock Universal language model fine-tuning for text classification.
\newblock In \emph{Proceedings of the 56th Annual Meeting of the Association for Computational Linguistics}.

\bibitem[{Hu et~al.(2021)Hu, Shen, Wallis, Allen-Zhu, Li, Wang, Wang, and Chen}]{lora}
Hu, E.~J.; Shen, Y.; Wallis, P.; Allen-Zhu, Z.; Li, Y.; Wang, S.; Wang, L.; and Chen, W. 2021.
\newblock Lora: Low-rank adaptation of large language models.
\newblock \emph{arXiv preprint arXiv:2106.09685}.

\bibitem[{Hu et~al.(2023)Hu, Lan, Wang, Xu, Lim, Lee, Bing, and Poria}]{llmadapter}
Hu, Z.; Lan, Y.; Wang, L.; Xu, W.; Lim, E.-P.; Lee, R. K.-W.; Bing, L.; and Poria, S. 2023.
\newblock LLM-Adapters: An Adapter Family for Parameter-Efficient Fine-Tuning of Large Language Models.
\newblock In \emph{Empirical Methods in Natural Language Processing}.

\bibitem[{Kaplun et~al.(2023)Kaplun, Gurevich, Swisa, David, Shalev-Shwartz, and Malach}]{kaplun2023less}
Kaplun, G.; Gurevich, A.; Swisa, T.; David, M.; Shalev-Shwartz, S.; and Malach, E. 2023.
\newblock Less is More: Selective Layer Finetuning with SubTuning.
\newblock \emph{arXiv preprint arXiv:2302.06354}.

\bibitem[{Koncel-Kedziorski et~al.(2016)Koncel-Kedziorski, Roy, Amini, Kushman, and Hajishirzi}]{mawps}
Koncel-Kedziorski, R.; Roy, S.; Amini, A.; Kushman, N.; and Hajishirzi, H. 2016.
\newblock {MAWPS}: A Math Word Problem Repository.
\newblock In \emph{Proceedings of NAACL}, 1152--1157.

\bibitem[{Lan et~al.(2020)Lan, Chen, Goodman, Gimpel, Sharma, and Soricut}]{layersharing}
Lan, Z.; Chen, M.; Goodman, S.; Gimpel, K.; Sharma, P.; and Soricut, R. 2020.
\newblock Albert: A lite bert for self-supervised learning of language representations.
\newblock In \emph{International Conference on Learning Representation}.

\bibitem[{Lee et~al.(2023)Lee, Chen, Tajwar, Kumar, Yao, Liang, and Finn}]{lee2022surgical}
Lee, Y.; Chen, A.~S.; Tajwar, F.; Kumar, A.; Yao, H.; Liang, P.; and Finn, C. 2023.
\newblock Surgical fine-tuning improves adaptation to distribution shifts.
\newblock In \emph{International Conference on Learning Representation}.

\bibitem[{Lei et~al.(2024)Lei, Bai, Brahma, Ainslie, Lee, Zhou, Du, Zhao, Wu, Li et~al.}]{lei2024conditional}
Lei, T.; Bai, J.; Brahma, S.; Ainslie, J.; Lee, K.; Zhou, Y.; Du, N.; Zhao, V.; Wu, Y.; Li, B.; et~al. 2024.
\newblock Conditional adapters: Parameter-efficient transfer learning with fast inference.
\newblock \emph{Advances in Neural Information Processing Systems}, 36.

\bibitem[{Lester, Al-Rfou, and Constant(2021)}]{prompttuning}
Lester, B.; Al-Rfou, R.; and Constant, N. 2021.
\newblock The power of scale for parameter-efficient prompt tuning.
\newblock In \emph{Empirical Methods in Natural Language Processing}.

\bibitem[{Li and Liang(2021)}]{prefix}
Li, X.~L.; and Liang, P. 2021.
\newblock Prefix-tuning: Optimizing continuous prompts for generation.
\newblock \emph{arXiv preprint arXiv:2101.00190}.

\bibitem[{Ling et~al.(2017)Ling, Yogatama, Dyer, and Blunsom}]{aqua}
Ling, W.; Yogatama, D.; Dyer, C.; and Blunsom, P. 2017.
\newblock Program Induction by Rationale Generation: Learning to Solve and Explain Algebraic Word Problems.
\newblock In \emph{Proceedings of the 55th Annual Meeting of the Association for Computational Linguistics (Volume 1: Long Papers)}, 158--167.

\bibitem[{Liu et~al.(2017)Liu, Simonyan, Vinyals, Fernando, and Kavukcuoglu}]{liu2017hierarchical}
Liu, H.; Simonyan, K.; Vinyals, O.; Fernando, C.; and Kavukcuoglu, K. 2017.
\newblock Hierarchical representations for efficient architecture search.
\newblock \emph{arXiv preprint arXiv:1711.00436}.

\bibitem[{Liu et~al.(2024)Liu, Wang, Yin, Molchanov, Wang, Cheng, and Chen}]{dora}
Liu, S.-Y.; Wang, C.-Y.; Yin, H.; Molchanov, P.; Wang, Y.-C.~F.; Cheng, K.-T.; and Chen, M.-H. 2024.
\newblock DoRA: Weight-Decomposed Low-Rank Adaptation.
\newblock In \emph{International Conference on Machine Learning}.

\bibitem[{Liu et~al.(2022)Liu, Ji, Fu, Tam, Du, Yang, and Tang}]{ptuning}
Liu, X.; Ji, K.; Fu, Y.; Tam, W.~L.; Du, Z.; Yang, Z.; and Tang, J. 2022.
\newblock P-tuning v2: Prompt tuning can be comparable to fine-tuning universally across scales and tasks.
\newblock In \emph{Proceedings of the 60th Annual Meeting of the Association of Computational Linguistics}.

\bibitem[{Mao et~al.(2021)Mao, Mathias, Hou, Almahairi, Ma, Han, tau Yih, and Khabsa}]{unipelt}
Mao, Y.; Mathias, L.; Hou, R.; Almahairi, A.; Ma, H.; Han, J.; tau Yih, W.; and Khabsa, M. 2021.
\newblock UniPELT: A Unified Framework for Parameter-Efficient Language Model Tuning.
\newblock \emph{ArXiv}, abs/2110.07577.

\bibitem[{Merity et~al.(2016)Merity, Xiong, Bradbury, and Socher}]{wikitext}
Merity, S.; Xiong, C.; Bradbury, J.; and Socher, R. 2016.
\newblock Pointer sentinel mixture models.
\newblock \emph{arXiv preprint arXiv:1609.07843}.

\bibitem[{Mihaylov et~al.(2018)Mihaylov, Clark, Khot, and Sabharwal}]{openbookqa}
Mihaylov, T.; Clark, P.; Khot, T.; and Sabharwal, A. 2018.
\newblock Can a suit of armor conduct electricity? a new dataset for open book question answering.

\bibitem[{Muennighoff et~al.(2022)Muennighoff, Wang, Sutawika, Roberts, Biderman, Scao, Bari, Shen, Yong, Schoelkopf et~al.}]{bloom}
Muennighoff, N.; Wang, T.; Sutawika, L.; Roberts, A.; Biderman, S.; Scao, T.~L.; Bari, M.~S.; Shen, S.; Yong, Z.-X.; Schoelkopf, H.; et~al. 2022.
\newblock Crosslingual generalization through multitask finetuning.
\newblock \emph{arXiv preprint arXiv:2211.01786}.

\bibitem[{Pan et~al.(2024)Pan, Liu, Diao, Pi, Zhang, Han, and Zhang}]{pan2024lisa}
Pan, R.; Liu, X.; Diao, S.; Pi, R.; Zhang, J.; Han, C.; and Zhang, T. 2024.
\newblock LISA: Layerwise Importance Sampling for Memory-Efficient Large Language Model Fine-Tuning.
\newblock \emph{arXiv preprint arXiv:2403.17919}.

\bibitem[{Patel, Bhattamishra, and Goyal(2021)}]{svamp}
Patel, A.; Bhattamishra, S.; and Goyal, N. 2021.
\newblock Are {NLP} Models really able to Solve Simple Math Word Problems?
\newblock In \emph{Proceedings of NAACL}, 2080--2094.

\bibitem[{Peng et~al.(2023)Peng, Li, He, Galley, and Gao}]{alpaca}
Peng, B.; Li, C.; He, P.; Galley, M.; and Gao, J. 2023.
\newblock Instruction Tuning with GPT-4.
\newblock \emph{arXiv preprint arXiv:2304.03277}.

\bibitem[{Pham et~al.(2018)Pham, Guan, Zoph, Le, and Dean}]{conf/icml/PhamGZLD18}
Pham, H.; Guan, M.~Y.; Zoph, B.; Le, Q.~V.; and Dean, J. 2018.
\newblock Efficient Neural Architecture Search via Parameter Sharing.
\newblock In \emph{International Conference on Machine Learning (ICML)}, 4092--4101.

\bibitem[{Ren et~al.(2021)Ren, Rajbhandari, Aminabadi, Ruwase, Yang, Zhang, Li, and He}]{offloading}
Ren, J.; Rajbhandari, S.; Aminabadi, R.~Y.; Ruwase, O.; Yang, S.; Zhang, M.; Li, D.; and He, Y. 2021.
\newblock ZeRO-Offload: Democratizing Billion-Scale Model Training.
\newblock \emph{arXiv preprint arXiv:2101.06840}.

\bibitem[{Sajjad et~al.(2023)Sajjad, Dalvi, Durrani, and Nakov}]{layerdrop}
Sajjad, H.; Dalvi, F.; Durrani, N.; and Nakov, P. 2023.
\newblock On the effect of dropping layers of pre-trained transformer models.
\newblock \emph{Computer Speech \& Language}, 77: 101429.

\bibitem[{Sakaguchi et~al.(2021)Sakaguchi, Bras, Bhagavatula, and Choi}]{winogrande}
Sakaguchi, K.; Bras, R.~L.; Bhagavatula, C.; and Choi, Y. 2021.
\newblock Winogrande: An adversarial winograd schema challenge at scale.
\newblock \emph{Communications of the ACM}, 64(9): 99--106.

\bibitem[{Sap et~al.(2019)Sap, Rashkin, Chen, LeBras, and Choi}]{siqa}
Sap, M.; Rashkin, H.; Chen, D.; LeBras, R.; and Choi, Y. 2019.
\newblock Socialiqa: Commonsense reasoning about social interactions.
\newblock \emph{arXiv preprint arXiv:1904.09728}.

\bibitem[{Sung, Cho, and Bansal(2022)}]{ladder_side_tuning}
Sung, Y.-L.; Cho, J.; and Bansal, M. 2022.
\newblock LST: Ladder Side-Tuning for Parameter and Memory Efficient Transfer Learning.
\newblock \emph{ArXiv}, abs/2206.06522.

\bibitem[{Touvron et~al.(2023)Touvron, Lavril, Izacard, Martinet, Lachaux, Lacroix, Rozi{\`e}re, Goyal, Hambro, Azhar et~al.}]{llama}
Touvron, H.; Lavril, T.; Izacard, G.; Martinet, X.; Lachaux, M.-A.; Lacroix, T.; Rozi{\`e}re, B.; Goyal, N.; Hambro, E.; Azhar, F.; et~al. 2023.
\newblock Llama: Open and efficient foundation language models.
\newblock \emph{arXiv preprint arXiv:2302.13971}.

\bibitem[{Wang and Komatsuzaki(2021)}]{gpt-j}
Wang, B.; and Komatsuzaki, A. 2021.
\newblock {GPT-J-6B: A 6 Billion Parameter Autoregressive Language Model}.
\newblock \url{https://github.com/kingoflolz/mesh-transformer-jax}.

\bibitem[{Wang et~al.(2022)Wang, Mukherjee, Liu, Gao, Awadallah, and Gao}]{wang2022adamix}
Wang, Y.; Mukherjee, S.; Liu, X.; Gao, J.; Awadallah, A.~H.; and Gao, J. 2022.
\newblock Adamix: Mixture-of-adapter for parameter-efficient tuning of large language models.
\newblock \emph{arXiv preprint arXiv:2205.12410}, 1(2): 4.

\bibitem[{Wei et~al.(2022)Wei, Wang, Schuurmans, Bosma, Xia, Chi, Le, Zhou et~al.}]{wei2022chain}
Wei, J.; Wang, X.; Schuurmans, D.; Bosma, M.; Xia, F.; Chi, E.; Le, Q.~V.; Zhou, D.; et~al. 2022.
\newblock Chain-of-thought prompting elicits reasoning in large language models.
\newblock \emph{Advances in neural information processing systems}, 35: 24824--24837.

\bibitem[{Zellers et~al.(2019)Zellers, Holtzman, Bisk, Farhadi, and Choi}]{hellaswag}
Zellers, R.; Holtzman, A.; Bisk, Y.; Farhadi, A.; and Choi, Y. 2019.
\newblock Hellaswag: Can a machine really finish your sentence?
\newblock \emph{arXiv preprint arXiv:1905.07830}.

\bibitem[{Zhang et~al.(2023{\natexlab{a}})Zhang, Ding, Qi, Zhu, Long, and Zhou}]{crash}
Zhang, K.; Ding, N.; Qi, B.; Zhu, X.; Long, X.; and Zhou, B. 2023{\natexlab{a}}.
\newblock CRaSh: Clustering, Removing, and Sharing Enhance Fine-tuning without Full Large Language Model.
\newblock In \emph{Empirical Methods in Natural Language Processing}.

\bibitem[{Zhang and He(2020)}]{zhang2020accelerating}
Zhang, M.; and He, Y. 2020.
\newblock Accelerating training of transformer-based language models with progressive layer dropping.
\newblock \emph{Advances in Neural Information Processing Systems}, 33: 14011--14023.

\bibitem[{Zhang et~al.(2023{\natexlab{b}})Zhang, Chen, Bukharin, He, Cheng, Chen, and Zhao}]{zhang2023adaptive}
Zhang, Q.; Chen, M.; Bukharin, A.; He, P.; Cheng, Y.; Chen, W.; and Zhao, T. 2023{\natexlab{b}}.
\newblock Adaptive budget allocation for parameter-efficient fine-tuning.
\newblock In \emph{International Conference on Learning Representations}. Openreview.

\bibitem[{Zhang et~al.(2024)Zhang, Qiang, Somayajula, and Xie}]{zhang2024autolora}
Zhang, R.; Qiang, R.; Somayajula, S.~A.; and Xie, P. 2024.
\newblock AutoLoRA: Automatically Tuning Matrix Ranks in Low-Rank Adaptation Based on Meta Learning.
\newblock \emph{arXiv preprint arXiv:2403.09113}.

\bibitem[{Zhang et~al.(2023{\natexlab{c}})Zhang, Roller, Goyal, Artetxe, Chen, Chen, Dewan, Diab, Li, Lin et~al.}]{opt}
Zhang, S.; Roller, S.; Goyal, N.; Artetxe, M.; Chen, M.; Chen, S.; Dewan, C.; Diab, M.; Li, X.; Lin, X.~V.; et~al. 2023{\natexlab{c}}.
\newblock Opt: Open pre-trained transformer language models.
\newblock \emph{URL https://arxiv.org/abs/2205.01068}, 3: 19--0.

\bibitem[{Zhu et~al.(2024)Zhu, Hu, Lin, and Han}]{zhu2023lift}
Zhu, L.; Hu, L.; Lin, J.; and Han, S. 2024.
\newblock LIFT: Efficient Layer-wise Fine-tuning for Large Model Models.

\bibitem[{Zhuang et~al.(2024)Zhuang, Yu, Wang, Sun, and Zhang}]{zhuang2024toolqa}
Zhuang, Y.; Yu, Y.; Wang, K.; Sun, H.; and Zhang, C. 2024.
\newblock Toolqa: A dataset for llm question answering with external tools.
\newblock \emph{Advances in Neural Information Processing Systems}, 36.

\bibitem[{Zoph and Le(2017)}]{conf/iclr/ZophL17}
Zoph, B.; and Le, Q.~V. 2017.
\newblock Neural Architecture Search with Reinforcement Learning.
\newblock In \emph{International Conference on Learning Representations (ICLR)}.

\end{thebibliography}

%%%%%%%%%%%%%%%%%%%%%%%%%%%%%%%%%%%%%%%%%%%%%%%%%%%%%%%%%%%%%%%%%%%%%%%%%%%%%%%
%%%%%%%%%%%%%%%%%%%%%%%%%%%%%%%%%%%%%%%%%%%%%%%%%%%%%%%%%%%%%%%%%%%%%%%%%%%%%%%
% APPENDIX
%%%%%%%%%%%%%%%%%%%%%%%%%%%%%%%%%%%%%%%%%%%%%%%%%%%%%%%%%%%%%%%%%%%%%%%%%%%%%%%
%%%%%%%%%%%%%%%%%%%%%%%%%%%%%%%%%%%%%%%%%%%%%%%%%%%%%%%%%%%%%%%%%%%%%%%%%%%%%%%

\clearpage
\appendix
\section{Appendix}
\subsection{Code and Reproducibility}
% \SetAlFnt{\fontsize{8pt}{9pt}\selectfont}
% \SetAlCapFnt{\fontsize{8pt}{9pt}\selectfont}
\begin{algorithm}[!h]
    %\SetAlgoLined
    \PyCode{import peft, transformers} \\
    \PyCode{from ist import IST} \\
    \PyComment{initialize the pre-trained model and PEFT modules} \\
    \PyCode{peft\_model = get\_peft\_model()} \\
    \PyComment{initialize IST as callback function} \\
    \PyCode{ist\_callback = IST()} \\
    \PyComment{adopt IST in Trainer with one modification} \\
    \PyCode{trainer = transformers.Trainer(model=peft\_model, callbacks=[ist\_callback]))}\\
    \PyCode{trainer.fit()}\\
    %\Indm %
    \caption{\fontsize{8pt}{9pt}\selectfont{IST, PyTorch-like}}
    \label{alg:code_box}
\end{algorithm}

Our code is based on the LLM-Adapter library\footnote{\href{https://github.com/AGI-Edgerunners/LLM-Adapters}{https://github.com/AGI-Edgerunners/LLM-Adapters}}~\cite{llmadapter}, a benchmark library for parameter-efficient fine-tuning (PEFT). To facilitate reproducibility, we have included the code, along with training scripts and instructions, in the supplementary material. Notably, our IST method is orthogonal to most PEFT methods and can be readily incorporated into the training process. As demonstrated in \autoref{alg:code_box}, our method requires only a single line of modification to the trainer based on the Hugging Face Transformers library\footnote{\href{https://github.com/huggingface/transformers}{https://github.com/huggingface/transformers}} and Peft library\footnote{\href{https://github.com/huggingface/peft}{https://github.com/huggingface/peft}}. Please refer to the code for more details.

\subsection{PEFT Overview}

\begin{table}[!h]
\centering
\vspace{2.8mm}
%\vspace{-5pt}
\setlength{\tabcolsep}{1pt}
\resizebox{0.48\textwidth}{!}{
\begin{tabular}
{l|cccc}
\hline
Method & Prompt & Repara & Series & Parallel  \\\hline
Prompt Tuning \cite{prompttuning}      &$\surd$ & & & \\
Prefix-Tuning  \cite{prefix}    & $\surd$& & & \\
% Spot   \cite{spot}            & $\surd$& & & \\
% IPT    \cite{ipt}        &$\surd$ & & & \\
LoRA     \cite{lora}       & &$\surd$ & & \\
KronA   \cite{krona}       & &$\surd$ & & \\
DoRA   \cite{dora}       & &$\surd$ & & \\
Adapters \cite{series}  & & &$\surd$ &        \\
AdaMix \cite{wang2022adamix}  & & &$\surd$ & \\
SparseAdapter \cite{sparseadapter}       & & &$\surd$ & \\
LeTS    \cite{lets}      & & &$\surd$ & \\
Parallel Adapter \cite{parallel} & & & & $\surd$ \\
LST \cite{ladder_side_tuning} & & & & $\surd$ \\
MAM Adapter  \cite{mam}    &$\surd$ &$\surd$ &$\surd$ & \\
UniPELT   \cite{unipelt}      &$\surd$ & $\surd$ &$\surd$ & \\
Compacter  \cite{compacter}     & &$\surd$ &$\surd$ & \\
S4-model  \cite{s4_model}     &$\surd$ &$\surd$ & & \\
\hline
\end{tabular}
}
\vspace{-2pt}
\caption{The PEFT methods are categorized based on the four common basic methods. "Prompt" represents prompt-based learning methods, "Repara" denotes reparametrization-based methods, "Series" is Series Adapters, and "Parallel" represents Parallel Adapters.
% \sj{Could fit it in single column format to save some space.}
}
\label{tab:peft_category}
\vspace{-8pt}
\end{table}

According to \citet{llmadapter} and \citet{peft_survey}, existing parameter-efficient fine-tuning (PEFT) methods can be roughly categorised into four types as shown in \autoref{tab:peft_category}.
In the following, we provide a brief overview of three layer-based PEFT methods used in our study: reparametrization-based methods, series adapters, and parallel adapters.

\begin{figure}[!t]
\centering
\includegraphics[width=.9\columnwidth]{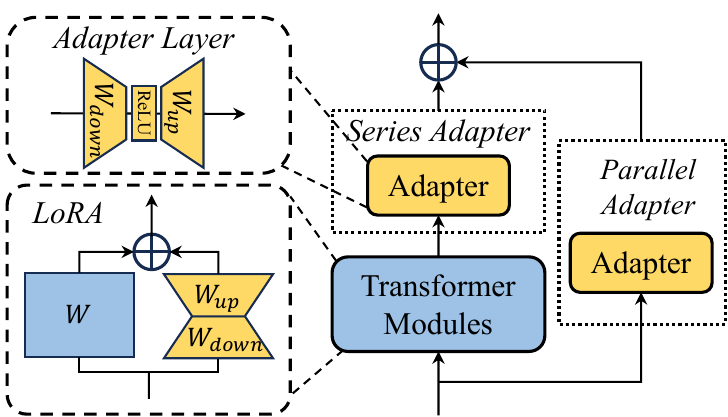}
\caption{Most existing PEFT approaches employ a layer-based design, consistently adding learnable modules or parameters to each layer of the transformer modules, including the Multi-Head Self-Attention (MHSA) and Feed-Forward Network (FFN).} 
\label{fig:peft} 
\end{figure}
\paragraph{Parallel Adapters.} Parallel adapters focus on incorporating additional learnable modules in parallel with distinct sublayers within the backbone model. The Parallel Adapter can be formulated as follows:
\begin{equation}
H_o \rightarrow H_o +f(H_i W_{down})W_{up}.
\end{equation}

\paragraph{Reparametrization-based method.}
This type of method aims to transform network weights using a low-rank technique. We take LoRA~\cite{lora} as an example of Reparametrization-based learning, which can be formulated below:
\begin{equation}
H_o = H_i W_0+H_i\Delta W = H_iW_0+H_iBA,
\end{equation}
where $H_i$ and $H_o$ are the input and output of a sublayer module (e.g., Linear), $W_0\in \mathbb{R}^{d\times d}$ can be any linear weight in the pre-trained LLM, $B \in \mathbb{R}^{r\times d}$ and $B \in \mathbb{R}^{d\times r}$ are lower-rank learnable matrix to approximate $\Delta W$. $r \ll d$ is the pre-defined rank for LoRA.

\begin{table}[!h]
\centering
% \vspace{2.8mm}
\vspace{-5pt}
\setlength{\tabcolsep}{3pt}
\begin{tabular}
{lccr}
\toprule
Dataset  & \# Train & \# Test &Answer  \\\midrule
{\textbf{Commonsense}}        & 170K & -    &- \\
\quad{BoolQ}          & 9.4K &3,270    &Yes/No \\
\quad{PIQA}          & 16.1K&1,830    &Option \\
\quad{SIQA}         & 33.4K&1,954    &Option \\
\quad{HellaSwag}    & 39.9K&10,042   &Option \\
\quad{WinoGrande}     & 63.2K&1,267    &Option \\
\quad{ARC-e}          & 1.1K &2,376    &Option \\
\quad{ARC-c}         & 2.3K &1,172    &Option \\
\quad{OBQA}           & 5.0K &500      &Option \\\midrule
{\textbf{Math10K}}      & 10K & -    &- \\
\quad{GSM8K}        & 8.8K &1,319    &Number \\
\quad{AQuA}         & 100K &254      &Option \\
\quad{MAWPS}      & -    &238      &Number \\
\quad{SVAMP}         & -    &1,000    &Number \\
\bottomrule
\end{tabular}
\vspace{-2pt}
\caption{The statistics of datasets for evaluation. \#~Train and \#~Test denote the number of training and test samples respectively.}
\label{tab:dataset_description}
%\vspace{-8pt}
\end{table}

\begin{table*}[!t]
\centering
\small
\begin{tabular}{c|ccccc|c}\toprule
\multirow{2}{*}{\begin{tabular}[c]{@{}c@{}}\textbf{Hyperparameters}\\ \textbf{(LoRA + IST)}\end{tabular}} & \multicolumn{5}{c|}{\textbf{Commonsense Reasoning}} & \textbf{Arithmetic Reasoning} \\
 & LLaMA$_{\text{7B}}$ & LLaMA$_{\text{13B}}$ & GPT-J$_{\text{6B}}$ & BLOOMz$_{\text{7B}}$ & LLAMA3$_{\text{8B}}$ & LLAMA3$_{\text{8B}}$ \\\midrule
Rank r & \multicolumn{5}{c|}{32} & 32 \\
$\alpha$ & \multicolumn{5}{c|}{64} & 64 \\
Dropout & \multicolumn{5}{c|}{0.05} & 0.05 \\
Optimizer & \multicolumn{5}{c|}{AdamW} & AdamW \\
LR & \multicolumn{5}{c|}{2e-4} & 1e-4 \\
LR Scheduler & \multicolumn{5}{c|}{Warmup Steps} & Warmup Steps \\
Batch size & \multicolumn{5}{c|}{16} & 16 \\
Warmup Steps & \multicolumn{5}{c|}{100} & 100 \\
Epochs & \multicolumn{5}{c|}{3} & 3 \\
Where & \multicolumn{5}{c|}{\{Q, K, V, Up, Down\}} & \{Q, K, V, Up, Down\} \\\hline
$\mu$ & \multicolumn{5}{c|}{10} & 100 \\
$\beta$ & \multicolumn{5}{c|}{ 0.25} &  0.25 \\
$N_L$, $N_u$, $N_v$ & {\{32, 8, 16\}} &  {\{40, 10, 20\}} & {\{28, 7, 14\}}   & {\{30, 8, 15\}}  &  {\{32, 8, 16\}} &  {\{32, 8, 16\}}\\
\bottomrule
\end{tabular}
\vspace{-2pt}
\caption{Hyperparameter configurations of IST for LLaMA-7B/13B, GPT-J 6B, BLOOMz 7B, and LLaMA3-8B with LoRA.}
\label{tab:hyper1}
%\vspace{-8pt}
\end{table*}

\begin{table}[!h]
\centering
\resizebox{0.49\textwidth}{!}{
\begin{tabular}{c|cccc}\toprule
\multirow{2}{*}{\textbf{Hyperparameters}} & \multicolumn{2}{c}{\textbf{Series Adapter + IST}} & \multicolumn{2}{c}{\textbf{Parallel Adapter + IST}} \\
 & LLaMA$_{\text{7B}}$ & LLaMA$_{\text{13B}}$ & LLaMA$_{\text{7B}}$ & LLaMA$_{\text{13B}}$\\\midrule
Bottleneck Size & \multicolumn{4}{c}{256} \\
Optimizer & \multicolumn{4}{c}{AdamW} \\
LR & \multicolumn{4}{c}{2e-4} \\
LR Scheduler & \multicolumn{4}{c}{Warmup Steps} \\
Batch size & \multicolumn{4}{c}{16} \\
Warmup Steps & \multicolumn{4}{c}{100} \\
Epochs & \multicolumn{4}{c}{3} \\
Where & \multicolumn{2}{c}{\{Up, Down\}} & \multicolumn{2}{c}{\{Up, Gate\}}  \\\hline
$\mu$ & \multicolumn{4}{c}{10} \\
 $\beta$ & \multicolumn{4}{c}{0.25} \\
$N_L$, $N_u$, $N_v$ &   {\{32, 8, 16\}} &  {\{40, 10, 20\}} & {\{32, 8, 16\}} &  {\{40, 10, 20\}}  \\
\bottomrule
\end{tabular}}
\vspace{-2pt}
\caption{Hyperparameter configurations of IST for LLaMA-7B/13B on commonsense reasoning tasks with series and parallel adapters.}
\label{tab:hyper2}
%\vspace{-8pt}
\end{table}

\paragraph{Series Adapters.} Series adapters involve incorporating additional learnable modules in a sequential manner within a specific sublayer. Series Adapter can be formulated as follows:
\begin{equation}
H_o \rightarrow H_o +f(H_o W_{down})W_{up},
\end{equation}
where $H_o$ is the output of a specific layer like MLP layer, $f(\cdot)$ is a non-linear function like ReLU, $W_{down}\in \mathbb{R}^{d\times r}$ and $ W_{up}\in \mathbb{R}^{r\times d}$ form a bottleneck MLP to save learnable parameters.

As shown in the \autoref{fig:peft}, the three PEFT methods mentioned above all utilize the layer-based design, i.e., adding identical learnable modules or parameters to each layer of the pre-trained LLM.

It is important to note that we did not include the prompt-based method in our comparison because original prompt tuning~\cite{prompttuning} is not a layer-based method; rather, it adds learnable soft prompts at the input layer. Furthermore, while some advancements in prompt tuning are layer-based, such as Prefix Tuning~\cite{prefix}, which independently adds soft prompts to the hidden states at all layers, they do not align with our design. This misalignment occurs because our proposed response suppression operates on the output of a PEFT method conditioned on the input, whereas prompt-based methods produce an output that is not conditioned on the input.

\subsection{Experimental Details}

\subsubsection{Dataset Statistics}
Detailed dataset statistics can be referred to \autoref{tab:dataset_description}. Note that we trained on Commonsense and Math10K for commonsense reasoning and arithmetic reasoning, respectively. During testing, we evaluated the predefined test sets of each dataset.

\subsubsection{Hyperparameters}
Detailed hyperparameter settings are provided in \autoref{tab:hyper1} and \autoref{tab:hyper2}. For PEFT training, we adhere to the settings outlined by LLM-Adapter Library~\cite{llmadapter}, with the exception of the learning rate. For IST, we consistently set $N_u$ to 25\% of the layers for the fine-tuning loop, $N_v$ to 50\% of the layers for the importance updating loop, and $\beta$ to 0.25. Additionally, $\mu$ is set to 10 and 100 for the Commonsense Reasoning task and the Arithmetic Reasoning task, respectively. 

\subsection{Additional Experiments}
\subsubsection{Importance Updating Rate $\mu$}
\begin{table}[t]
\setlength{\tabcolsep}{1.2mm}
\centering
\begin{tabular}{cccccc}\toprule
\multirow{2}{*}{Random} & \multicolumn{5}{c}{IST} \\ 
 & $\mu$=0.1 & $\mu$=1 & $\mu$=10 & $\mu$=100 & $\mu$=1000 \\\hline
75.8 & 75.7 & 75.9 & \textbf{76.5} & 74.8 & 73.9\\
\bottomrule
\end{tabular}
\vspace{-2pt}
\caption{Sensitivity of importance updating rate $\mu$.}
\label{tab:mu}
%\vspace{-8pt}
\end{table}

The updating rate of importance is associated with several hyperparameters, such as $T_c$, $N_c$, $N_v$, and $\mu$. To narrow the hyperparameter search space and reduce the complexity of using IST, we fixed most hyperparameters, setting $T_c$ to 10, $N_c$ to 3, and $N_v$ to half the number of layers. We then adjusted the importance updating rate $\mu$ to match with the dynamics of PEFT fine-tuning. This parameter is largely dependent on the maximum number of training iterations. If $\mu$ is too small, the method will approximate a random strategy. Conversely, if $\mu$ is too large, the method will tend to train only a fixed set of layers. As shown in \autoref{tab:mu}, we consider $\mu$ values of [0.1, 1, 10, 100, 1000]. $\mu$ is a parameter that exhibits insensitivity, indicating the robustness of our method.

\begin{table}[t]
\centering
\begin{tabular}{lc}\toprule
Method & Results \\\midrule
Baseline & 74.7 \\
IST with $\beta = 0$ & 75.0 \\
IST with $\beta = 0.1$ & 76.3 \\
IST with $\beta = 0.25$ & \textbf{76.5} \\
IST with $\beta = 0.5$ & 75.3\\
\bottomrule
\end{tabular}
\vspace{-2pt}
\caption{Effect of response suppression factor $\beta$ within IST for LLaMA-7B on commonsense reasoning tasks with LoRA.}
\label{tab:rsf}
\vspace{-8pt}
\end{table}

\begin{table*}[t]
% \vspace{0.8em}
\setlength{\tabcolsep}{1.2mm}
\centering
\resizebox{0.99\textwidth}{!}{
\begin{tabular}{clcccccccccc}
\toprule
\textbf{Model} & \multicolumn{1}{c}{\textbf{PEFT}}  & \textbf{BoolQ} & \textbf{PIQA}&\textbf{SIQA}& \textbf{HellaSwag} & \textbf{WinoGrande}& \textbf{ARC-e} & \textbf{ARC-c} & \textbf{OBQA} & \textbf{Avg.} \\ \hline
ChatGPT&\multicolumn{1}{c}{-}&73.1&85.4&68.5&78.5&66.1&89.8&79.9&74.8&77.0\\\hline
\multirow{2}{*}{LLaMA3$_{\text{8B}}$}&DoRA$_{ICML2024}$&74.6&89.3&79.9&95.5&85.6&90.5&80.4&85.8&\textbf{85.2}\\
&DoRA + IST&74.0&89.2&80.2&95.0&86.2&90.3&81.2&85.6&\textbf{85.2}\\
\bottomrule
\end{tabular}}
\vspace{-2pt}
\caption{Adaptability to the latest LoRA-variant method called DoRA~\cite{dora}. Our approach can reduce memory consumption without compromising accuracy.}
\label{tab:dora}
\vspace{-8pt}
\end{table*}

\subsubsection{Response Suppression Factor $\beta$}
\autoref{tab:rsf} illustrates the effect of varying the response suppression factor $\beta$ on training a LLaMA 7B model using the commonsense dataset. We evaluated among four values: [0, 0.1, 0.25, 0.5]. When the factor is set to 0, which is equivalent to dropping the PEFT modules within the layer, it does not adequately reflect the importance of PEFT. Increasing the factor enhances performance, peaking at 0.25. This indicates that compared to removing the PEFT modules, suppressing its output better captures its influence on the loss. However, further increasing the factor to 0.5 results in diminished effectiveness, likely due to reduced variation in loss. These findings suggest that a relatively small, non-zero factor is optimal for accurately estimating the PEFT module's impact on loss.

\subsubsection{Adaptability to SoTA PEFT method}
To demonstrate the versatility of the IST method, we integrated IST into a recent LoRA variant called DORA~\cite{dora}, which decouples the low-rank component into direction and magnitude, yielding better performance. As shown in \autoref{tab:dora}, our method can enhance the DoRA method in Commonsense Reasoning tasks without any loss of performance, while also requiring less memory and computational resources. This efficiency is achieved by explicitly training only a subset of all layers, highlighting the general applicability of our proposed IST.

% \subsubsection{Configuration of $N_v$}
\subsection{Time Consumption}
\begin{table}[!h]
\centering
\small
\begin{tabular}{c|ccc}\toprule
 & \textbf{LoRA} & \textbf{LISA} & \textbf{LoRA + IST} \\\midrule
\begin{tabular}[c]{@{}c@{}}Forward time \\ per iter. (ms)\end{tabular} & 135 & \textbf{101} & 135 \\\hline
\begin{tabular}[c]{@{}c@{}}Backward time \\ per iter. (ms)\end{tabular} & 184 & 225 & \textbf{150} \\\hline
\begin{tabular}[c]{@{}c@{}}Time consumption \\ per 100 iter. (s)\end{tabular} & \textbf{31.9} & 32.6 & 32.6\\
\bottomrule
\end{tabular}
\vspace{-2pt}
\caption{Comparison of Training Times. All results were obtained using one Nvidia GTX 4090 GPU.}
\label{tab:time}
\vspace{-8pt}
\end{table}
To accurately estimate the training time, we randomly sampled prompts from the Alpaca dataset and limited the maximum output token length to 1024. We used LoRA on LLaMA-7B with a rank of 32 as our baseline. Additionally, we employed the LISA~\cite{pan2024lisa} method, which randomly selects two transformer layers for updating. We conducted 140 iterations and averaged the forward and backward times of the middle 100 iterations to obtain a stable time estimate during training. As shown in \autoref{tab:time}, LISA reduces the forward time compared to LoRA due to the absence of additional parameters for inference, while it increases the backward time. Conversely, IST maintains the forward pass time but reduces the backward time by approximately 10\%. Despite this, IST requires an additional three forward passes every 10 fine-tuning loops for importance updating. Consequently, after 100 iterations, the total time consumption for IST becomes comparable to that of LISA and slightly higher than LoRA.

\end{document}